\documentclass[lettersize,journal]{IEEEtran}
\usepackage{amsmath,amsfonts}
\usepackage{algorithmic}
\usepackage{algorithm}
\usepackage{array}
\usepackage[caption=false,font=normalsize,labelfont=sf,textfont=sf]{subfig}
\usepackage{textcomp}
\usepackage{stfloats}
\usepackage{url}
\usepackage{verbatim}
\usepackage{graphicx}
\usepackage{cite}
\usepackage{amssymb}
\usepackage{float}
\usepackage{afterpage}
\usepackage{placeins}
\usepackage{hyperref}
\usepackage{cleveref}
\crefname{figure}{fig}{figures}
\Crefname{figure}{Fig}{Figures}
\usepackage{multirow}
\usepackage[normalem]{ulem}
\useunder{\uline}{\ul}{}
\usepackage{amssymb}
\usepackage{bbding}
\usepackage{pifont}
\usepackage{wasysym}
\usepackage{utfsym}
\usepackage{fontawesome}
\usepackage{colortbl}
\usepackage{threeparttable}
\usepackage{bm}

\begin{document}

\title{Targetless Intrinsics and Extrinsic Calibration of Multiple LiDARs and Cameras with IMU using Continuous-Time Estimation}

\author{Yuezhang Lv$^{1}$, Yunzhou Zhang*$^{1}$, Chao Lu$^{2}$, Jiajun Zhu$^{2}$, Song Wu $^{1}$ 
\thanks{*The corresponding author of this paper}
\thanks{$^{1}$Yuezhang Lv, Yunzhou Zhang are with College of Information Science and Engineering, Northeastern University, Shenyang 110819, China
        {\tt\small zhangyunzhou@mail.neu.edu.cn}}%
\thanks{$^{2}$Chao Lu, Jiajun Zhu work at MEGVII Technology, China}%
\thanks{This work was supported by National Natural Science Foundation of China (No. 61973066), Major Science and Technology Projects of Liaoning Province(No. 2021JH1/10400049),   Fundamental Research Funds for the Central Universities(N2004022).}%
}



\maketitle

\begin{abstract}

Accurate spatiotemporal calibration is a prerequisite for multisensor fusion. However, sensors are typically asynchronous, and there is no overlap between the fields of view of cameras and LiDARs, posing challenges for intrinsic and extrinsic parameter calibration. To address this, we propose a calibration pipeline based on continuous-time and bundle adjustment (BA) capable of simultaneous intrinsic and extrinsic calibration (6 DOF transformation and time offset). We do not require overlapping fields of view or any calibration board. Firstly, we establish data associations between cameras using Structure from Motion (SFM) and perform self-calibration of camera intrinsics. Then, we establish data associations between LiDARs through adaptive voxel map construction, optimizing for extrinsic calibration within the map. Finally, by matching features between the intensity projection of LiDAR maps and camera images, we conduct joint optimization for intrinsic and extrinsic parameters. This pipeline functions in texture-rich structured environments, allowing simultaneous calibration of any number of cameras and LiDARs without the need for intricate sensor synchronization triggers. Experimental results demonstrate our method's ability to fulfill co-visibility and motion constraints between sensors without accumulating errors.

\end{abstract}


\section{Introduction}\label{section:1}
Autonomous vehicles and robots are often equipped with multiple sensors such as LiDARs, cameras, and IMUs to perform perception tasks like SLAM, object detection, and tracking. LiDAR directly measures depth information but lacks detailed environmental perception; cameras provide rich texture information but are sensitive to lighting conditions; IMUs offer high-frequency acceleration and angular velocity information, unaffected by external environmental factors, allowing different sensors to complement each other. With the advancement of BEV (Bird's-Eye View) perception accuracy, there is widespread attention on multisensor fusion solutions.

One key step in multi-sensor fusion involves estimating the 6 degrees of freedom (DoF) transformation - rotation, translation, and time offset between individual sensors, known as extrinsic calibration of sensors. Many calibration methods rely on infrastructure (such as calibration boards) to aid in calibration \cite{kalibr}\cite{mishra2020experimental}\cite{dhall2017lidar}, yet the high cost limits the practicality of real-world tasks. In contrast, natural scene calibration is closer to the actual application environment. Richer environmental information can enhance the diversity of sensor data features, allowing for calibration in a wide range of environments and conditions. Sensors used in practical applications may not be perfectly synchronized, hence necessitating time calibration to compute the time offset between sensors.

The current mainstream calibration methods for LiDAR-IMU, camera-IMU, and LiDAR-camera pairs involve calibrating individual sensors, inevitably leading to accumulated errors. Joint calibration of sensors becomes highly necessary. The traditional discrete-time representation in sensor fusion with asynchronous measurements results in an increase in the number of optimization variables as the sensor count rises, leading to higher computational costs. \cite{oa-licalib} introduced a continuous-time calibration method. The advantage of continuous-time representation is that any sensor can interpolate its corresponding pose on splines based on timestamps without introducing new optimization variables, enabling elegant joint observation of asynchronous sensors. Concerning establishing shared observations among sensors, the sparsity of LiDAR data makes it challenging to create common observations among LiDARs. \cite{mlcc} noted that some methods overlook the cumulative errors in LiDAR odometry during calibration, hence proposed a planar Bundle Adjustment-based data association approach to minimize odometry errors. Inspired by this, we extend its application to the calibration of multiple LiDARs-IMU.


Therefore, we have developed a continuous-time-based Multi-Camera-LiDAR-IMU (M-LIC) calibration method without the use of infrastructure to compute the intrinsic and extrinsic (spatial and temporal) parameters of LIC. Our method builds upon camera self-calibration methods \cite{colmap}, which recover intrinsic parameters, camera poses, and visual points through SFM, constructing common observations between cameras. Additionally, we introduce a continuous-time based LiDAR plane BA method that associates spatial structural information between LiDAR frames and establishes common observations between LiDAR-LiDAR frames, simultaneously reducing LiDAR odometry errors while calibrating LiDAR-IMU extrinsics and time offsets. Furthermore, we introduce a method based on aligning LiDAR intensity images with camera images for cross-modal camera-LiDAR association and depth estimation for visual points. By jointly calibrating parameters using motion information and common observations from each sensor, we mitigate accumulated calibration errors from individual sensors. In summary, our contributions are as follows:

\begin{itemize}

\item We present a novel continuous-time-based calibration method that enables the simultaneous estimation of intrinsic, extrinsic parameters, and time offsets for multiple LiDAR-camera-IMU systems.
\item We propose an innovative and robust targetless calibration method that fully utilizes structural information and texture features in natural scenes with motion information from IMUs.
\item We propose a multi-cameras data association method based on SFM, and a multi-LiDARs data association method based on voxel map, which do not require overlapping FOVs among cameras and LiDAR.
\item We conduct experiments by collecting real-world data, and the experimental results show that our method can calibrate all sensors simultaneously with high accuracy.

\end{itemize}

\section{Related Works}\label{section:2}

Calibration is mainly divided into target calibration and calibration in natural scenes, here we mainly introduce the more practical calibration method in natural scenes.

\subsection{LiDAR-IMU calibration}

For sensors such as LiDAR-IMUs that do not have co-vision, the coordinate transformation between two frames is usually estimated based on the motion of the rigid body.

Some LiDAR-inertial odometry systems have a built-in initialization process, but these initialization modules are usually simple and incomplete, like \cite{fast-lio2}\cite{lio-sam}\cite{lins}, these algorithms do not take into account time offsets due to data transfer and require several seconds of standstill time for the calibrated gyroscope bias.Since when the device is kept stationary, the gravity and accelerometer bias are coupled, the accelerometer bias is not calibrated during its initialization. The above mentioned odometry require good initial values of the external bias although they can be calibrated online during operation.

Some existing calibration methods designed for LiDAR-IMUs have higher accuracy, e.g., \cite{lidar_IMU_calib}\cite{oa-licalib} proposed a continuous-time based calibration method that models the trajectory of the IMU with the external parameters as a problem of solving the spline optimal control points. \cite{Li-init} proposes a robust, fast and real-time initialization method for LIDAR IMU, which can effectively calibrate the state of the external parameters and time difference between LIDAR and IMU, as well as the gyroscope and accelerometer offsets. However, the above methods do not take into account the odometry error of the LiDAR and are based on the frame-to-map alignment method, which is difficult to be extended to the joint calibration of multiple LiDARs. \cite{BALM}\cite{BALM2.0} proposes to dynamically divide the map into voxels of different sizes, and the voxel maps correlate the co-vision features between the LiDAR frames, which reduces the cumulative odometer error by optimizing the map. Based on this \cite{mlcc} proposed a voxel map based multi-LiDAR calibration method for LiDARs with small FOV, but the algorithm did not consider the time offset between the calibration of the LiDAR and IMU and the LiDAR.

\subsection{Camera-IMU calibration}

In 2012, Furgale et al. proposed the Kalibr algorithm \cite{kalibr}, which formulates the continuous-time SLAM problem as a maximum likelihood estimation of B-spline control points using B-spline interpolation. This approach reduces the number of optimized states. Subsequent work on continuous-time calibration mostly follows this framework. Building upon this foundation, Furgale et al. subsequently explored the time offset between calibrating cameras and IMUs \cite{furgale2013unified} and the shutter timing of rolling shutter cameras \cite{oth2013rolling}.

For multi-camera-IMU calibration, \cite{heng2013camodocal} runs monocular visual odometry for each camera and performs extrinsic hand-eye calibration based on pose without establishing inter-camera co-visibility. Ensuring calibration accuracy, establishing co-visibility between cameras is crucial. Li and Heng et al. \cite{li2013multiple} proposed a descriptor-based calibration method and Matlab toolbox, requiring adjacent cameras to simultaneously observe a part of the calibration pattern. With the advancements in SLAM and SFM, Heng et al. \cite{heng2015leveraging} \cite{Multi-Camera-Rig} utilized costly SLAM methods to construct high-precision maps of calibration regions. Once the map construction is completed, multi-camera calibration is performed through image-based localization, akin to using a known calibration board. These methods do not necessitate overlapping fields of view between cameras nor initial extrinsic parameters.

\subsection{LiDAR-Camera calibration}

LiDAR-camera calibration can be categorized into motion-based and motionless methods. In motion-based methods, initial extrinsic parameters are usually obtained from hand-eye calibration and refined by the co-vision information of the LiDAR and the image. In motionless methods, only the features that are co-visualized in the fields of view of the two sensors are extracted and matched, and then the extrinsic parameters are optimized by minimizing the reprojection or maximizing the mutual information between the point and the image.

LiDAR-camera cross-modal associations are usually constructed in two ways, one is nearest neighbor search by visually recovered 3D feature points in 3D space \cite{joint_hku}, but this method does not take into account the error of visual 2D feature point detection, and due to the existence of the error of visual feature point detection, the accuracy of depth recovery of the feature points is planarly and positively correlated with the square of the ranging, which usually leads to false associations; the other is to perform the correlation by the ray projection method for correlating SemLoc \cite{semloc} and then minimizing the visual points to the map plane, but most of the visual feature points are on the line. Another is to project the LiDAR points being projected into the image space for cross-modal alignment. \cite{livox_camera_calib} \cite{cocalibration} \cite{zhou2024targetless} extracts line features in the space, but the method relies on a structured scene, the LiDAR-visual line feature mismatch rate is larger, and most of the texture information of the surface in the environment is discarded in this method; \cite{calib-anything} obtains the final outer reference by segmenting the point cloud, objects in the image, and by matching the overlap between objects, but it is subject to the segmentation algorithm's accuracy However, it is limited by the accuracy of the segmentation algorithm, and the calibration accuracy is poor. \cite{direct_visual_lidar_calibration} projects the LiDAR point cloud to the image plane, and the intensity of the point cloud is used as the gray value of the image, and cross-modal correlation is carried out by the alignment method of deep learning, and we believe that the reliability and accuracy of this method is the best, and pixel-level calibration can be achieved, but this method usually requires multiple scenes to be jointly calibrated in order to ensure the accuracy.

\section{PROBLEM FORMULA-TION}
\label{section:3}
To facilitate a clear understanding of the subsequent analysis, we first introduce the notation used throughout this paper. We use ${ }_A^B T \in SE(3)$  to denote the 6-DoF transformation.${ }_A^B T=\left[\begin{array}{cc}{ }_A^B R & { }^B p_A \\ 0 & 1\end{array}\right]$ consists of a rotation part ${ }_A^B R \in SO(3)$  and a translation part ${ }^B p_A \in \mathbb{R}^3$. For simplicity, we omit the homogeneous conversion in the rigid transformations of ${ }^B p={ }_A^B T^A p$. In this paper, we assume that LiDAR $\{L\}$, camera $\{C\}$ and IMU $\{I\}$ are rigidly connected. At the beginning of calibration, define the coordinate system of the IMU in the first frame as the world coordinate system $\{G\}$. Define the time stamp of the lidar as the reference time, and the time offset between the lidar and the IMU is defined as $t d_I$, and the time offset between the lidar and the camera is defined as $t d_c$. The parameters calibrated in this paper include: rotation ${ }_L^I R$, translation ${ }^I p_L$ and time offset from LiDAR frame to IMU frame $t d_I$, rotation ${ }_C^I R$, translation ${ }^I p_C$ and time offset from camera frame to IMU frame $t d_c$, camera intrinsic $\{D\}$ and IMU intrinsic $x_I$.

\begin{equation}
\chi=\left\{x_p, x_q, x_I,{ }_I^L R,{ }^L p_I, t d_I,{ }_I^C R,{ }^C p_I, t d_C, D,{ }^G p_c,{ }^G \varepsilon\right\}
\end{equation}

Where the control points $\left\{x_p, x_q\right\}$ of the B-spline represent the continuous-time trajectory, see \ref{section:3-1} below for derivation. Landmark of the camera is defined as ${ }^G p_c$, and the LiDAR plane landmark is defined as ${ }^G \varepsilon$.

Based on the collected LiDAR, camera, and IMU data, we can construct the following nonlinear least squares (NLS) problem:

\begin{equation}
\hat{\chi}=\underset{\chi}{\arg \min }\left(r_I+r_C+r_L+r_{lc}\right)
\end{equation}

where $r_L$, $r_C$, $r_I$, $r_{lc}$ are the measurement residual functions for the LiDAR, the camera, the IMU, the LiDAR-to-camera, respectively, which will be derived in detail below;

\subsection{Continuous-Time Trajectory Representation}
\label{section:3-1}
The B-spline curve provides analytic derivatives in closed form, which facilitates the integration of asynchronous measurements for state estimation. The B-spline curve also has the property of being locally controllable, which means that the update of a single control point affects only a few consecutive segments of the spline curve, a property that limits the number of control points. In the calibration algorithm of this paper, we use B-splines to parameterize the trajectory of 6-degree-of-freedom IMU in the world coordinate system $\{G\}$. In order to decouple the rotation and translation, we use two uniform B-splines to represent the 3-degree-of-freedom translations and rotations, respectively. The position can be calculated by interpolating the timestamps measured by the sensors into each of the two splines.

The B-spline is determined by the order $k$, $N+1$ control points $\{ x_0,x_1,\ldots,x_N\}$, $M$ node vectors $\{ t_0,t_1,\ldots,t_M\}$, and $M=N+k+1$. The B-spline is defined as follows:

\begin{equation}
x(t)=\sum_{i=0}^NB_{i,k}(t)x_i
\end{equation}

\begin{equation}
x(t)=\tilde{B}_{i, k}(t) x_0+\sum_{i=0}^N \tilde{B}_{i, k}(t) d_i
\end{equation}

where $\tilde{B}_{i, k}(t)=\sum_{s=i}^N B_{i, k}(t) d_i, \quad d_i=x_i-x_{i-1}$, and the coefficients of the uniform B-Spline are constants that can be written in matrix form \cite{sommer2020efficient}. This matrix form can also be used for cumulative B-splines. To simplify the calculation, the B-Spline sampling equation at time $t$ is changed to:

\begin{equation}
x(u)=x_i+\sum_{j=1}^{k-1} \lambda_i(u) \cdot d_j^i
\end{equation}

where $u=\left(t-t_i\right) /\left(t_{i+1}-t_i\right), t \in\left[t_i-t_{i+1}\right]$, the coefficients $\lambda_i(u)$ are constant and depend only on the order of the B-spline, and the difference $d_j^i=x_{i+j}-x_{i+j-1}$.

Extending the above equation to Lie groups, while $\lambda_i(u)$  can be used to scale the vector $d_j^i$ in a translation, there is no notion of scaling for an element $R$ of a Lie group. Therefore, we must first map $R$ from the Lie group to the Lie algebra space, which is a vector space, then scale it, and finally map it back under the Lie group space: $E X P(\lambda \cdot \log (R))$, with the cumulative B-splines of $SO(3)$ defined as:

\begin{equation}
R(u)=R_i \cdot \sum_{j=1}^{k-1} \operatorname{Exp}\left(\lambda_i(u) \cdot d_j^i\right)=R_i \cdot \sum_{j=1}^{k-1} A_i
\end{equation}

where $d_j^i=\log \left(R_{i+j-1}^{-1} \cdot R_{i+j}\right)$, $A_i=\operatorname{Exp}\left(\lambda_i(u) \cdot d_j^i\right)$.

\subsection{IMU residuals}

In this paper, we use the third-order B-spline to parameterize the trajectory of IMU, and since the B-spline provides analytic derivatives in closed form, the derivatives of the spline concerning time can be easily computed:

\begin{equation} \label{equ:spline derivative}
\begin{aligned}
& { }^G a(t)={ }^G \ddot{p}_I(t)=\sum_{j=1}^3 \ddot{\lambda}_j(t) \cdot d_j^i \\
& { }_I^G \dot{R}(t)=R_i\left(\dot{\mathrm{A}}_1 \mathrm{~A}_2 \mathrm{~A}_3+\mathrm{A}_1 \dot{\mathrm{A}}_2 \mathrm{~A}_3+\mathrm{A}_1 \mathrm{~A}_2 \dot{\mathrm{A}}_3\right)
\end{aligned}
\end{equation}

Assuming an angular velocity ${ }^I \omega_m$ and a linear acceleration ${ }^I a_m$ for the raw IMU measurements at ${ }t_I$, the following equation holds based on the angular velocity and linear acceleration in Eq. \ref{equ:spline derivative}:

\begin{equation}
\begin{aligned}
& { }^I \omega_m={ }_I^G R^T(t){ }^I \omega(t)+b_\omega \\
& { }^I a_m={ }_I^G R^T(t)\left({ }^I a(t)-{ }^G g\right)+b_a
\end{aligned}
\end{equation}

where, $b_\omega$ $b_a$ is the IMU bias. The residuals of the IMU can be defined as follows:

\begin{equation} \label{equ:r_I}
\begin{aligned}
& r_\omega={ }^I \omega_m-{ }_I^G R^T(t){ }^I \omega(t)-b_\omega \\
& r_a={ }^I a_m-{ }_I^G R^T(t)\left({ }^I a(t)-{ }^G g\right)-b_a
\end{aligned}
\end{equation}

\subsection{LiDAR plane BA Factor}
\label{section:3-3}
In order to construct data association between different LiDAR and intra-LiDAR, inspired by \cite{BALM}\cite{BALM2.0},\cite{HBA} we implement a multi-LiDAR data association method based on planar BA. Assuming that the LiDAR's position is known, for each scan of the LiDAR point cloud, we downsample the current scan, extract the planner features of the point cloud by curvature judgment \cite{loam}, and construct a point cloud map based on the position of each scan. We suppose that the point cloud map is composed of many planes of different sizes, so we cut the whole point cloud map into many voxels of different sizes according to a preset size, then determine whether the point cloud within each voxel can be fitted into a plane, if it can form a plane, it is considered as a planar voxel; vice versa, we continue to split this voxel into eight small voxels, and then check whether they are planar voxels, and if not, repeat the above operation until the minimum lower limit of voxel size is reached. As shown in the figure, with these voxels, the spatial structure can be utilized to construct data associations for the LiDAR, and these planes are referred to as signposts.

To prevent excessive over-parameterization of the plane within the voxel, we uses the closest point approach\cite{lips} to realize the 3 DOF parameterization of the plane features, the closest point parameterization method is expressed by combining the normal vector of the plane with the distance $d$ of the closest point of the plane from the origin of the coordinate system to the origin of the coordinate system, and the closest point plane parameterization method is denoted as:

\begin{figure}
    \vspace{0cm}
    \setlength{\abovecaptionskip}{0.1cm} 
    \centering
    \includegraphics[width = 8.5cm]{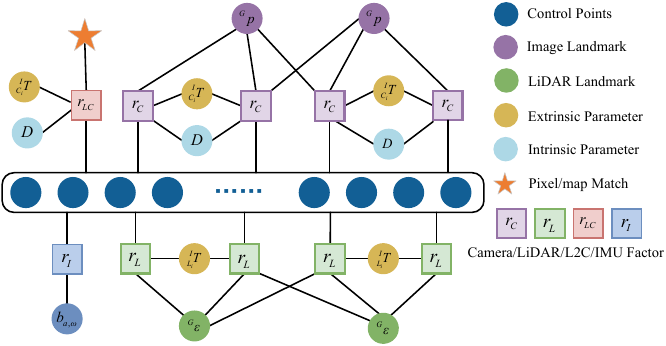}
    \caption{Factor graphs. We achieve joint optimization of continuous-time trajectort, extrinsic and intrinsic parameters by combining visual measurements, LiDAR measurements, IMU measurements and LiDAR-visual measurements}
    \label{fig:1}
\end{figure}

\begin{equation}
\varepsilon=n d
\end{equation}

Assuming a total of $m$ LiDAR and $h$ voxels in the voxel map denoted as $V_j$ ($j \in(0, h]$), each voxel contains $w$ distinct LiDAR points ${ }^{L_i} p_k$ ($i \in(0, m]$, $k \in(0, w]$). The plane BA residual factors of the LiDAR can be constructed as follows:

\begin{equation} \label{equ:r_L}
r_L\left(t_p,{ }^{L_i} p, \varepsilon\right)=\frac{{ }^G \varepsilon}{\left\|{ }^G \varepsilon\right\|} \cdot{ }_I^{L_i} T \cdot T\left(t_L\right) \cdot{ }^L p+\left\|{ }^G \varepsilon\right\|
\end{equation}

where ${ }^{L_i} p$ is the original point of the LiDAR scan, $t_L$ is the timestamp of the LiDAR point, ${ }^G \varepsilon_j$ represents the parametric equation defining the plane in the world coordinate system corresponding to the respective voxel.

\begin{figure*}
    \setlength{\abovecaptionskip}{-0.1cm} 
    \centering
    \includegraphics[width = 14.5cm]{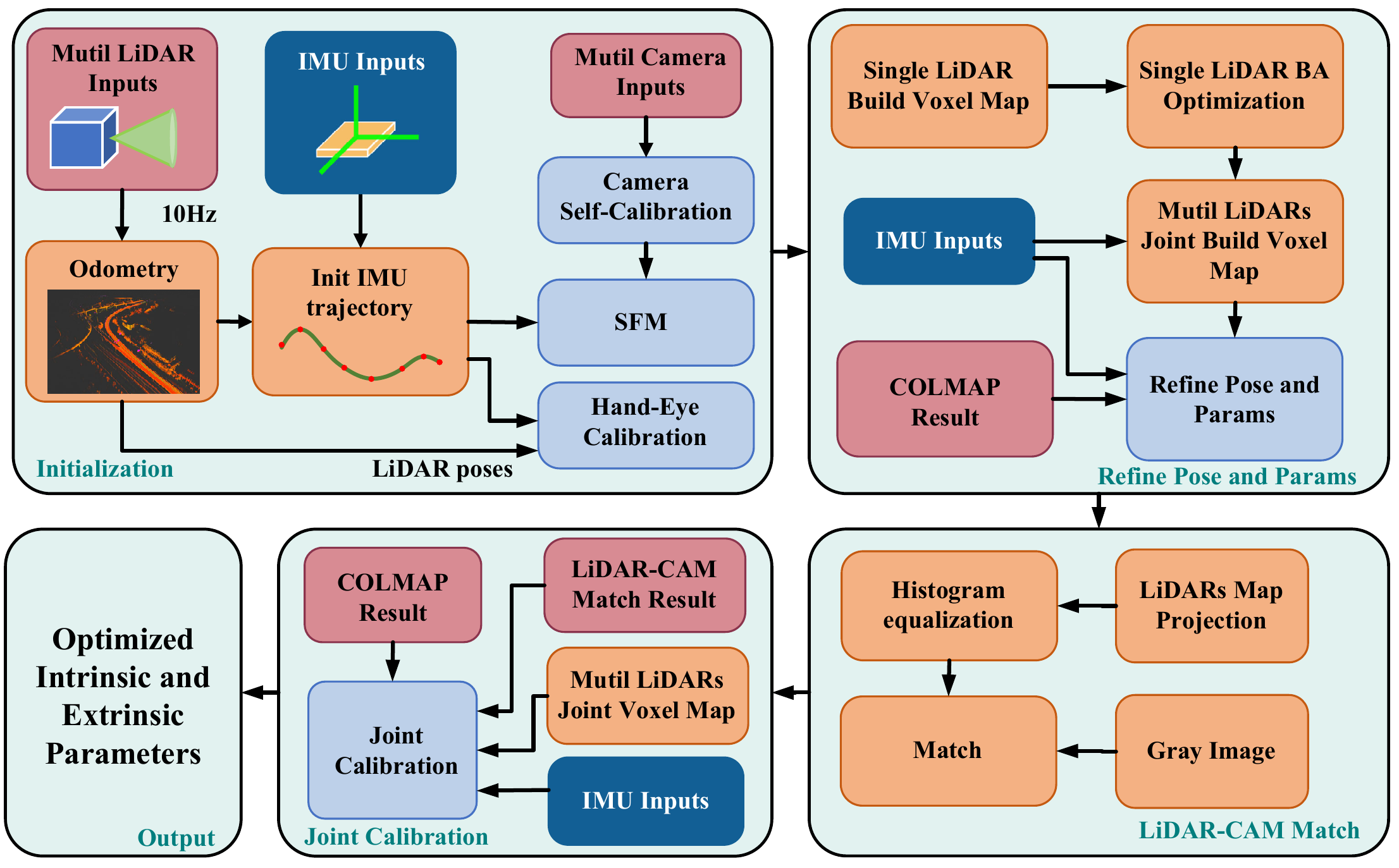}
    \caption{System Pipeline. Each block is explained in the corresponding subsection of \ref{section:4} with explicit reference back to this diagram.}
    \label{fig:overview}
\end{figure*}

\subsection{Camera BA Factor}
\label{sectio:3-4}
Compared with LiDAR, camera can easily construct common observations among frames through feature point and their descriptors \cite{orb-slam3}, in order to facilitate the calibration of the intrinsic parameters, we use 3D vectors to represent the camera landmark, assuming that a landmark ${ }^G p$ under the world coordinate system is observed by the image with timestamp $t_c$, let the point in the image pixel coordinates as $u v$, the following equation can be constructed:

\begin{equation} \label{equ:r_c}
r_c\left({ }_c^I T, t d_c, D,{ }^G p\right)=\pi_c\left({ }_{C_i}^I T \cdot T\left(t_c+t d_c\right) \cdot{ }^G p_j, D\right)-u v_k
\end{equation}

\subsection{LiDAR-camera Factor}
\label{sectio:3-5}

Suppose there are $w$ 3D feature points in the world coordinate system is ${ }^G p$, each observed across $h$ image frames. The pixel coordinates of the 2D feature points in the observed images are denoted as $u v_k$, where $k$ ranges within $\left(0, h\right]$. For each 2D feature point, the residual can be constructed by the camera projection model:

\begin{equation} \label{equ:r_l_c}
r_{l c}(D)=\pi_c\left({ }_c^I T \cdot T\left(t_c+t d_c\right) \cdot{ }^G p, D\right)-u v_k
\end{equation}

\section{METHODOLOGY}
\label{section:4}
In our approach, we assume the presence of $m$ cameras and $n$ LiDARs. The map coordinate system for LiDAR odometry is denoted as $M$. The function $u v=\pi_C\left({ }^C p, D\right)$ represents the camera projection, transforming a 3D point ${ }^C p$ in space into a 2D pixel $u v \in \mathbb{R}^2$ on the image plane, where $D$ is the camera intrinsic parameter, including focal length, optical center, and aberration coefficient.

Fig. \ref{fig:overview} illustrates our system diagram. It mainly contains the following parts: (\ref{section:4-1}) we use the LiDAR to assist the initialization of the IMU continuous-time trajectory, while the multi-camera performs the calibration of the camera's intrinsic parameters through COLMAP, then initializes the LiDAR-IMU extrinsic parameters according to the hand-eye calibration, and re-performs the sfm according to the result of COLMAP to initialize the camera-IMU extrinsic parameters; (\ref{section:4-2}) before constructing the cross-LiDAR co-observation using voxel map, we carry out the refinement of single LiDAR BA and align it to the IMU trajectory, after which the joint cross-LiDAR and cross-camera co-observations are optimized with the motion information of the IMU; (\ref{section:4-3}) the optimized LiDARs joint global map is projected onto the cameras imaging plane based on the calibrated parameters and positions, and matched with the gray image to obtain 2D feature points with corresponding 3D map points; (\ref{section:4-4}) based on the observation of the LiDAR and the camera, the residuals of joint part \ref{section:4-2} are jointly optimized for the joint intrinsic and extrinsic parameters.

\subsection{Initialization}
\label{section:4-1}
\subsubsection{Spline control points initialization}
\label{section:4-1-1}
When initializing the IMU spline trajectory, relying solely on the original IMU measurements becomes unreliable due to the introduction of two zero elements in the vector $u$ of Eq. \eqref{equ:spline derivative} caused by the second derivative of the B-Spline \cite{lidar_IMU_calib}. Consequently, we need to initialize the spline with the help of the LiDAR's position data. In this process, one of the LiDARs is designated as the base LiDAR, establishing a continuous-time trajectory for its position. Leveraging the IMU's acceleration and angular velocity measurements alongside the continuous-time LiDAR trajectory, we construct the following residuals:

\begin{equation}
\begin{aligned}
r_a\left({ }_L^I R,{ }^I p_L\right) & ={ }_L^I R\left({ }_L^M R(t)-1\left({ }^L p(t)-{ }^G g\right)\right) \\
& +\left\lfloor{ }_L^M \ddot{R}(t)\right\rfloor{ }^I p_L+\left\lfloor{ }_L^M \dot{R}(t)\right\rfloor^2{ }^I p_L-{ }^I a_m \\
r_\omega\left({ }_L^I R,{ }^I p_L\right) & ={ }_L^I R{ }^L \omega(t)-{ }^I \omega_m
\end{aligned}
\end{equation}

where $\lfloor\cdot\rfloor$ denotes the antisymmetric matrix of vectors, ${ }^I \omega_m$ and ${ }^I a_m$ are the IMU's acceleration and angular velocity measurements.

From the raw IMU measurements, we can solve for the LiDAR-IMU extrinsic parameters ${ }_L^I R$ and ${ }^I p_L$ by addressing the following least squares problem:

\begin{equation}
{ }_L^I R,{ }^I p_L,{ }^G g=\arg \min \sum_k \frac{1}{2}\left\|r_{a, k}\right\|_{\Sigma_a}^2+\frac{1}{2}\left\|r_{\omega, k}\right\|_{\Sigma_\omega}^2
\end{equation}

The extrinsic parameters are obtained, and the LiDAR position can be converted to the IMU position using ${ }_L^I R$:

\begin{equation}
{ }_I^M T={ }_L^I T^{-1} \cdot{ }_L^M T
\end{equation}

Once the IMU's position is determined, we can align it with the world coordinate system using gravity, effectively initializing the continuous-time trajectory of the IMU.

\subsubsection{Intrinsic parameters initialization}

To calibrate the camera's intrinsic parameters in scenes without specific targets, we employ a rough calibration method using Structure from Motion (SfM). This involves utilizing the open source software COLMAP \cite{colmap}. To enhance the association of feature points across different viewpoints, we utilize superpoint for extracting image features and superglue for match. During SfM, the initial values of the camera's intrinsic parameters are derived from the image's dimensions and field of view. The aberration coefficient is initially set to zero. COLMAP then registers the images using vision Bundle Adjustment (BA), progressively estimating parameters for all frames. This includes intrinsic parameters, camera positions, and 3D points.

After registering all the frames, we acquire the camera positions, 2D image feature points and 3D feature points, and intrinsic parameters. However, the positions and 3D feature points are tied to the coordinate system of the first camera frame and lack scale. Therefore, we rely on the 2D image feature points and their associations, then we performe another SFM. This time, we incorporate the positions of the IMU to realign the 3D points under the world coordinate system. Further details on this process will be elucidated in the subsequent section.

\subsubsection{Extrinsic parameters, camera 3D point initialization}

For the LiDAR extrinsic parameters, we initialize the extrinsic parameters by aligning the position increments of the two sensors from LiDAR and IMU. Our method of choice for LiDAR odometry is the robust, generalized KISS ICP. Using the discrete positions derived from the LiDAR odometry, we can compute the LiDAR position ${ }_{L_k}^M T_i$, where $i$ ranges within $\left(0, N\right]$ and provides the relative position ${ }_{L_k}^{L_{k+1}} T_i$ between consecutive LiDAR frames.

Simultaneously, based on the continuous time trajectory of the IMU in section. \ref{section:4-1-1}, we can determine the relative position ${ }_{I{k+1}}^{I_k}T$ within the IMU frame for the time interval $\left[t_k, t_{k+1}\right]$. The relative position of the LiDAR and the IMU at $k$ should satisfy the following equation:

\begin{equation}
{ }_{I_{k+1}}^{I_k} T \cdot{ }_{L_i}^I T={ }_{L_i}^I T \cdot{ }_{L_{k+1}}^{L_k} T_i
\end{equation}

The above equation can be transferred to another equation representation:

\begin{equation}
\left\{\begin{array}{c}
\left(\mathcal{L}\left({ }_{I_{k+1}}^{{ }^{I_k}} q\right)-\mathcal{R}\left({ }_{L_{k+1}}^{L_k} q\right)\right){ }_L^I q=0 \\
\left({ }_{I_{k+1}}^{{ }_{L_k}} q-I_{3 \times 3}\right)^I p_L={ }_L^I q^{I_k} p_{I_{k+1}}-{ }^{I_k} p_{L_{k+1}}
\end{array}\right.
\end{equation}

Define the unit quaternions $q=\left[\begin{array}{ll}\mathbf{q}_v & q_w\end{array}\right]^T$, $\mathcal{L}(q)$ and $\mathcal{R}(q)$ to be multiplication matrices defined as follows:

\begin{equation}
\begin{gathered}
\mathcal{L}(\bar{q})=\left[\begin{array}{cc}
q_w \mathbf{I}_3-\left\lfloor\mathbf{q}_v\right\rfloor & \mathbf{q}_v \\
-\mathbf{q}_v^T & q_w
\end{array}\right] \\
\mathcal{R}(\bar{q})=\left[\begin{array}{cc}
q_w \mathbf{I}_3+\left\lfloor\mathbf{q}_v\right\rfloor & \mathbf{q}_v \\
-\mathbf{q}_v^T & q_w
\end{array}\right]
\end{gathered}
\end{equation}

Jointly multiframed, the form of the above equation can be written in the form of $\mathrm{A}_L^I q=0, \mathrm{~B}^I p_L=C$. Thus, the LiDARs can be solved for the extrinsic parameters with imu.

For extrinsic parameters of camera and visual 3D feature points, we utilizes solely the 2D image feature points and their corresponding associations to execute a recalibration of SFM. This process relies on the continuous-time trajectory data obtained from the IMU and the results generated by COLMAP. All visual feature points are united by the following equation:

\begin{equation}
{ }_{C_i}^I T,{ }^G p=\arg \min \sum_{i=0}^n \sum_{j=0}^w \sum_{k=0}^h \frac{1}{2}\left\|r_c\right\|^2
\end{equation}

By solving the above equations, we can estimate the extrinsic parameters of the camera-IMU, the 3D feature points in the world coordinate system. In this process, we do not need to optimize the control points of the IMU trajectory, so it is done quickly.

\subsection{Refine Pose and Extrinsic Params}
\label{section:4-2}

In the initialization phase, ensuring the robustness and speed of optimization involves relying on the presumed accuracy of the LiDAR-assisted IMU position. To enhance the precision of both position and extrinsic parameters, refinement occurs by iteratively adjusting spline control points and extrinsic parameters, leveraging observations from cameras, LiDARs, and IMU data.

\begin{figure}
    \vspace{0cm}
    \setlength{\abovecaptionskip}{0.1cm} 
    \centering
    \includegraphics[width = 8.5cm]{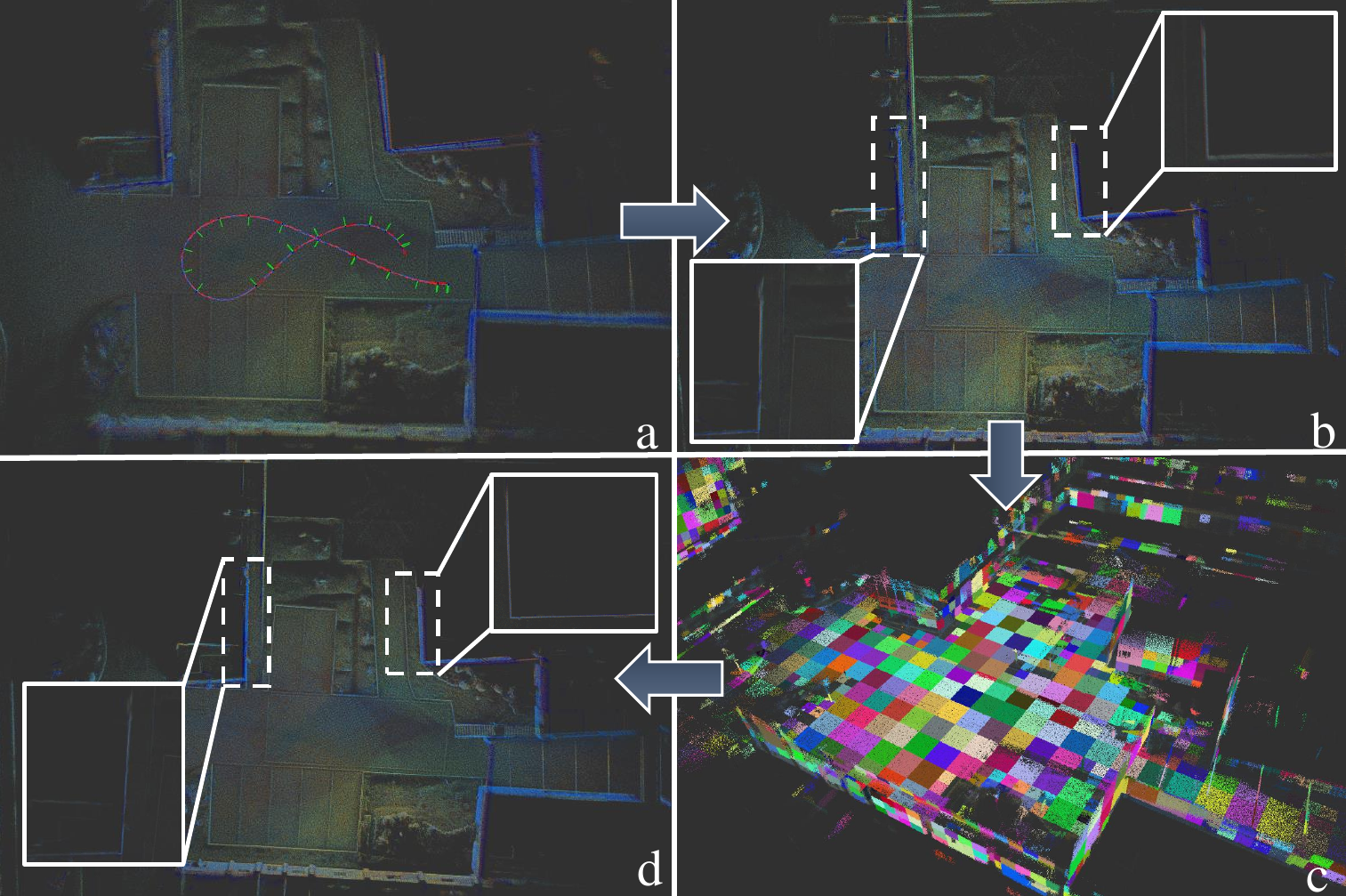}
    \caption{LiDAR BA calibration. a) Before calibration, initialization errors caused layering in the point cloud, making it challenging for multiple LiDARs to collectively establish a voxel map. b) After single LiDAR BA calibration, the consistency of LiDAR point cloud is enhanced, but there is still a little stratification. c) Multi-LiDARs co-construction of the voxel map. d) After calibration, the layering in the point cloud disappears, aligning it with the IMU trajectory, resulting in improved map consistency.}
    \label{fig:4}
\end{figure}

We leverage the voxel maps detailed in part \ref{section:3-3} to establish co-visibility relationships within and among LiDAR. The above part illustrates the covision relationship constructed by structural information, which relies on the consistency of the map. While individual LiDAR consistency is assured by KISS-ICP\cite{kiss-icp}. Unfortunately, it is difficult to ensure the consistency among LiDARs due to the existence of the error in part \ref{section:4-1} mentioned above, and there will be point cloud mismatches among the LiDARs as shown in Fig. \ref{fig:4} (a).  Therefore, in order to improve the global consistency of the maps, individual LiDAR Bundle Adjustment (BA) optimization is performed to refine the extrinsic parameters of the LiDAR-IMU before constructing the multi-LiDAR voxel maps. After the optimization is completed, the LiDAR point cloud map will be aligned to the IMU spline trajectory, and the point cloud consistency among different LiDARs will be guaranteed, as shown in Fig. \ref{fig:4} (d). 

Based on this, we can construct a voxel map of multiple LiDARs, and thus construct data association within and between LiDARs. Joint image measurements (Eq. \ref{equ:r_c}), LiDAR measurements (Eq. \ref{equ:r_L}), and IMU measurements (Eq. \ref{equ:r_I}), refine the continuous-time trajectory and extrinsic parameters.

\begin{figure}
    \vspace{0cm}
    \setlength{\abovecaptionskip}{0.1cm} 
    \centering
    \includegraphics[width = 8.5cm]{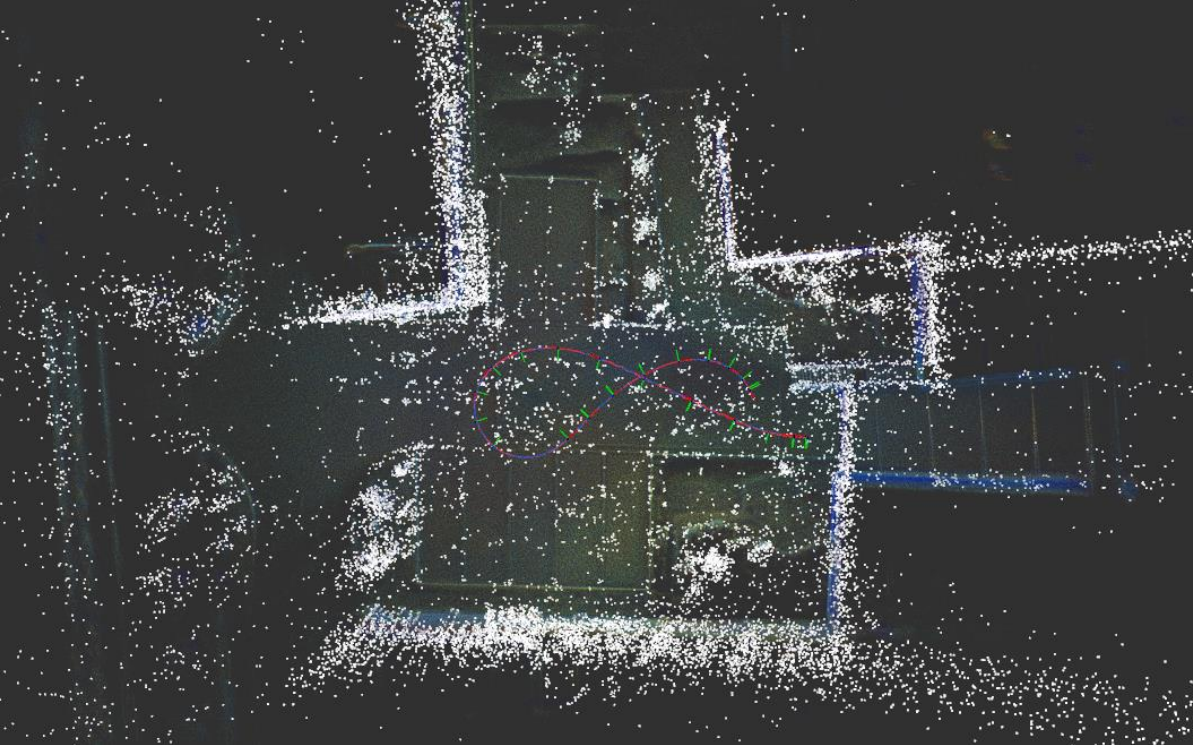}
    \caption{Pose and extrinsic parameters refined with LiDAR point cloud in purple and visual point cloud map in white.}
    \label{fig:lidar_image_map}
\end{figure}

The BA process will optimize the IMU trajectory and extrinsic parameters by optimizing the map, and the global consistency between the LiDAR point cloud map and the visual point cloud map will be better. As shown in the following Fig. \ref{fig:lidar_image_map}.

\subsection{LiDAR-Camera Match}
\label{section:4-3}
The above process completes the co-calibration between the LiDAR and the camera, and obtains the position and external parameters with higher consistency, but the LiDAR and the camera are indirectly associated with each other through the trajectory of the IMU, and there is no direct construction of the co-visualization between the LiDAR and the camera, in order to refine the camera's intrinsic parameters, and to obtain the extrinsic parameters with higher consistency, the correlation between the LiDAR and the camera is constructed in this section, and the co-optimization will be carried out on the basis of this relationship.

\begin{figure}
    \vspace{0cm}
    \setlength{\abovecaptionskip}{0.1cm} 
    \centering
    \includegraphics[width = 8.5cm]{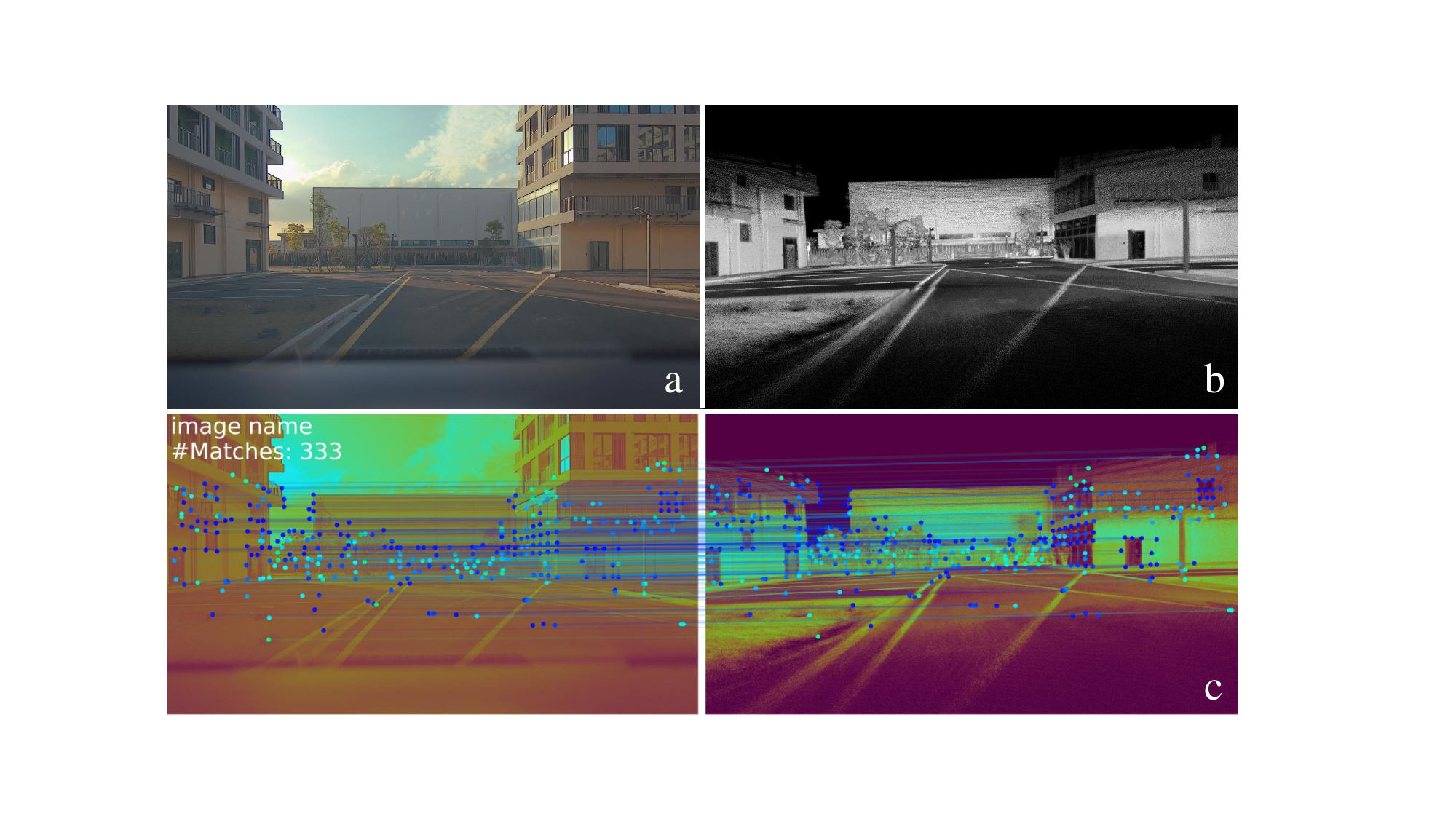}
    \caption{a) Original image. b) Projection and rendering of the LiDAR map onto the imaging plane based on camera pose. c) Matching results between the camera, original image, and intensity map.}
    \label{fig:lidar_image_match}
\end{figure}

First, we project the LiDAR point cloud in 3D space onto the camera's imaging plane based on the camera's position and the camera's imaging model, and generate an intensity image by taking the LiDAR's intensity as the gray value of the image, as well as grayscale the image, as shown in Fig. \ref{fig:lidar_image_match} (a). Although we just rendered each point without interpolation and gap filling, the rendered results exhibit good appearance quality due to the dense accumulated point cloud. When generating the intensity image, in order to ensure real-time performance, we do not project each frame of the camera, but select the frame with the largest parallax for projection based on the position sampling.

We then use surperpoint \cite{superpoint} and superglue \cite{superglue} to match the intensity image to the gray map, which gives us the 2D feature points on the image as well as the 3D points in the world coordinate system, as shown in Fig. \ref{fig:lidar_image_match} (c).

\subsection{Joint Calibration}
\label{section:4-4}

According to the matching results in Section \ref{section:4-3}, we can get the image feature points and their 3D points in the world coordinate system. After removing some outliers by pnp ransac, each point can be constructed as a residual by the camera projection model according to Eq. (\ref{equ:r_l_c}). Since the feature points have direct 3D point observations, the accuracy of the intrinsic parameters of the camera can be guaranteed, as in the case of the targeted calibration approach. The joint calibration of internal and extrinsic parameters is realized according to Eq. (\ref{equ:r_L}) (\ref{equ:r_c}) (\ref{equ:r_I}) (\ref{equ:r_l_c}). It is worth noting that in this optimization, we consider the LiDAR map to be optimal and there is no need to optimize the voxel plane. All the above nonlinear least squares problems are automatically derived by the ceres solver.

\section{EXPERIMENTS}

To evaluate the effectiveness of the calibration algorithm, we conduct calibration experiments of multi-LiDARs-Cameras-IMU under self-driving vehicles in this paper. We give the initial values of the calibrated extrinsic parameters as structural values, the initial rotation and translation errors of the extrinsic parameters are 4.81°/10 cm, and the time offset is 0ms. All the experiments are carried out on computers with an AMD Ryzen7 5800H CPU and a 16G RAM PC.

\begin{table*}[]
\small
\centering
\tabcolsep=0.1cm
\caption{Accuracy evaluation of the method in this paper on autonomous vehicles}
\label{table:accuracy_evaluation}
\renewcommand{\arraystretch}{1.4}
\begin{tabular}{cccccccccc}
\hline{}
\multirow{2}{*}{Method}      &                  & \multicolumn{3}{c}{\textit{p}}               & \multicolumn{3}{c}{\textit{q}}            & \multicolumn{2}{c}{RMSE} \\ \cline{2-10} 
                             &                  & px (m±cm)          & py (m±cm)          & pz (m±cm)               & roll (deg±deg)         & pitch (deg±deg)       & yaw (deg±deg)          & Pos(cm)    & Rot(deg)    \\ \hline
\multirow{4}{*}{lidar-imu}   & front\_lidar     & 0.012±1.44  & 1.693±2.01  & \textbackslash{} & 99.02±0.1    & 89.16±0.19  & 9.03±0.11    & 0.921      & 0.079       \\
                             & back\_lidar      & 0.007±-1.26 & 0.257±1.45  & \textbackslash{} & -90.16±0.1   & -4.83±0.20  & 179.2±0.12   & 0.657      & 0.083       \\
                             & left\_lidar      & -0.423±1.07 & 0.967±1.25  & \textbackslash{} & 179.54±0.14  & -6.31±0.12  & 179.81±0.17  & 0.884      & 0.079       \\
                             & right\_lidar     & 0.511±1.64  & 0.976±2.62  & \textbackslash{} & -0.86±0.1    & -4.99±0.13  & -179.46±0.17 & 1.167      & 0.079       \\ \hline
\multirow{4}{*}{cam-imu}     & cam\_back\_120   & 0.003±1.74  & -0.951±1.14 & \textbackslash{} & -179.96±0.12 & -0.39±0.11  & -87.82±0.19  & 0.745      & 0.085       \\
                             & cam\_left\_190   & -0.918±0.93 & 2.389±1.2   & \textbackslash{} & 130.25±0.07  & 0.45±0.12   & -92.7±0.18   & 0.636      & 0.084       \\
                             & cam\_right\_190  & 0.952±1.55  & 2.394±1.52  & \textbackslash{} & -129.6±0.19  & -0.56±0.1   & -92.24±0.08  & 0.729      & 0.082       \\
                             & cam\_front\_120  & 0.027±0.86  & 1.972±0.8   & \textbackslash{} & -0.28±0.12   & -0.19±0.11  & -90.79±0.17  & 0.431      & 0.079       \\ \hline
\multirow{4}{*}{lidar-lidar} & L\_C\_to\_L\_B     & 0.009±0.33  & -0.436±2.08 & -0.726±1.06      & -96.17±0.04  & 0.54±0.01   & -89.41±0.02  & 0.769      & 0.016       \\
                             & L\_D\_to\_L\_B     & -0.011±0.16 & 0.499±1.23  & -0.717±0.64      & 95.13±0.06   & -0.78±0.03  & 89.64±0.02   & 0.508      & 0.023       \\
                             & L\_C\_to\_L\_A     & -0.706±1.2  & -0.434±1.29 & -0.053±0.22      & -89.77±0.07  & -5.51±0.02  & 174.96±0.03  & 0.563      & 0.025       \\
                             & L\_D\_to\_L\_A     & -0.706±1.23 & -0.434±1.3  & -0.053±0.23      & 88.88±0.03   & -5.71±0.01  & -174.6±0.02  & 0.683      & 0.015       \\ \hline
\multirow{3}{*}{cam-cam}     & C\_D\_to\_C\_E     & 0.470±0.19  & -0.387±0.20 & -0.823±0.57      & -2.47±0.01   & -49.73±0.01 & -179.78±0.01 & 0.210      & 0.005       \\
                             & C\_D\_to\_C\_F     & -0.484±0.76 & -0.378±0.19 & -0.805±0.33      & 1.5±0.01     & 48.65±0.02  & -178.2±0.02  & 0.315      & 0.007       \\
                             & C\_E\_to\_C\_F     & -1.229±1.04 & -0.026±0.37 & -1.413±0.43      & 157.56±0.01  & 80.9±0.01   & -22.65±0.01  & 0.423      & 0.011       \\ \hline
\multirow{4}{*}{lidar-cam}   & L\_A\_to\_C\_A & -0.010±1.12 & -0.737±0.39 & -1.238±0.5       & 84.89±0.05   & 87.34±0.02  & 174.47±0.01  & 0.547      & 0.022       \\
                             & L\_B\_to\_C\_D & -0.015±0.59 & -0.392±0.1  & -0.273±1.08      & -90.33±0.02  & 1.62±0.02   & 179.72±0.08  & 0.492      & 0.034       \\
                             & L\_C\_to\_C\_E & -0.598±1.07 & -0.747±0.23 & -1.127±1.8       & 167.51±0.12  & 49.37±0.02  & -98.84±0.01  & 0.713      & 0.050       \\
                             & L\_D\_to\_C\_F & 0.579±1.15  & -0.785±0.05 & -1.083±1.15      & 8.58±0.09    & 47.63±0.02  & 98.38±0.01   & 0.695      & 0.030       \\ \hline
\end{tabular}
\end{table*}

\subsection{Autonomous Driving Dataset}

\begin{figure}
    \vspace{0cm}
    \setlength{\abovecaptionskip}{0.1cm} 
    \centering
    \includegraphics[width = 8.5cm]{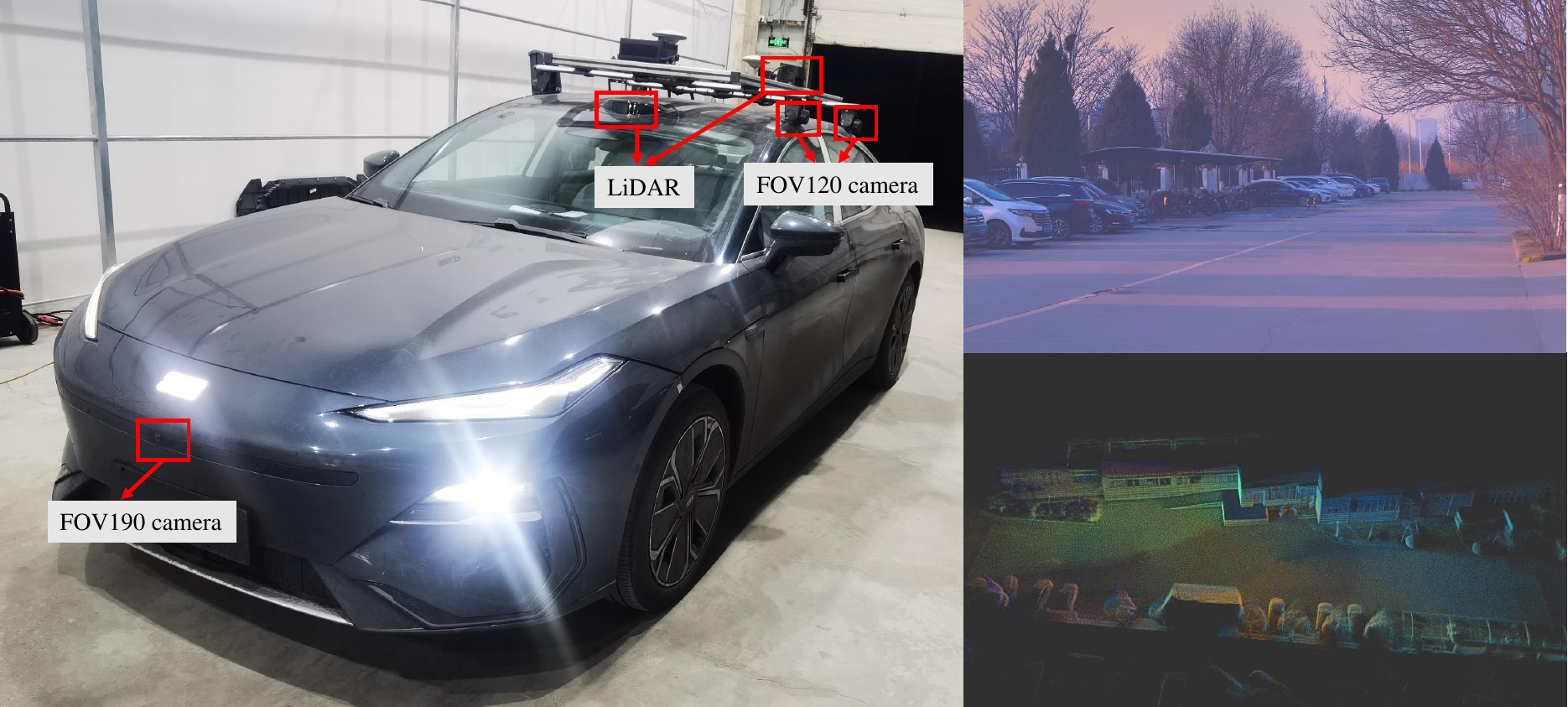}
    \caption{The self-driving vehicle is fitted with 4 solid-state LiDARs, 10 cameras, with high-precision inertial guidance, and this dataset is collected while in the park.}
    \label{fig:megvii_car}
\end{figure}

\begin{table}[]
\small
\centering
\tabcolsep=0.1cm
\caption{Sensor topic and name in this paper}
\label{table:sensor_topic}
\setlength{\tabcolsep}{8pt}
\renewcommand{\arraystretch}{1.2}
\begin{tabular}{cccc}
\hline
                        & Sensor topic    & Name  & FOV  \\ \hline
\multirow{4}{*}{LiDAR}  & back\_lidar     & L\_A & 120° \\
                        & front\_lidar    & L\_B & 120° \\
                        & left\_lidar     & L\_C & 120° \\
                        & right\_lidar    & L\_D & 120° \\ \hline
\multirow{6}{*}{Camera} & cam\_back\_120  & C\_A & 120° \\
                        & cam\_back\_190  & C\_B & 190° \\
                        & cam\_front\_190 & C\_C & 190° \\
                        & cam\_front\_120 & C\_D & 120° \\
                        & cam\_left\_190  & C\_E & 190° \\
                        & cam\_right\_190 & C\_F & 190° \\ \hline
INS                     & INS             & imu   &      \\ \hline
\end{tabular}
\end{table}

In order to verify the superiority of our method for multi-LiDARs-Camera-IMU calibration, we collect data for experiments on a self-built autopilot device, as shown in Fig. \ref{fig:megvii_car}, which includes four solid-state LiDARs with a horizontal FOV of 120 (30° area of common view between adjacent LiDAR), two Cameras with FOV 120 (for driving, including a pair of binocular cameras) with four fisheye cameras with FOV 190 (for parking), and a high-precision inertial guidance RTK. The sensor configuration is shown in Table \ref{table:sensor_topic}. For ease of representation, the sensor is renamed in this paper.

When collecting the dataset, the collector drove the car around the park in a figure of eight, and collected multiple sequences with a duration of 120 seconds, which were divided into six segments, each of which circled a complete figure of eight. Since the motion around the figure of eight is rotated along the z-axis of the vehicle, the external translation of the sensor-IMU along the axis of rotation is insignificant according to the analysis\cite{yang2019degenerate}, so we do not estimate the z-axis translation, and the translation of the structural values of the car is carefully designed at the factory with a small error.

The method presented in this paper can calibrate all sensor parameters in one go. Calibration was performed on these six sequences, and the results were tabulated, as shown in Table \ref{table:accuracy_evaluation}. It can be seen that, for sensors with a common view, the calibration accuracy of this paper is 0.04 degrees and 0.5 cm; for sensors without a common view, the calibration accuracy is 0.08 degrees and 1 cm.

\subsection{Camera Intrinsics Calibration Experiment}

\begin{figure}
    \vspace{0cm}
    \setlength{\abovecaptionskip}{0.1cm} 
    \centering
    \includegraphics[width = 8.5cm]{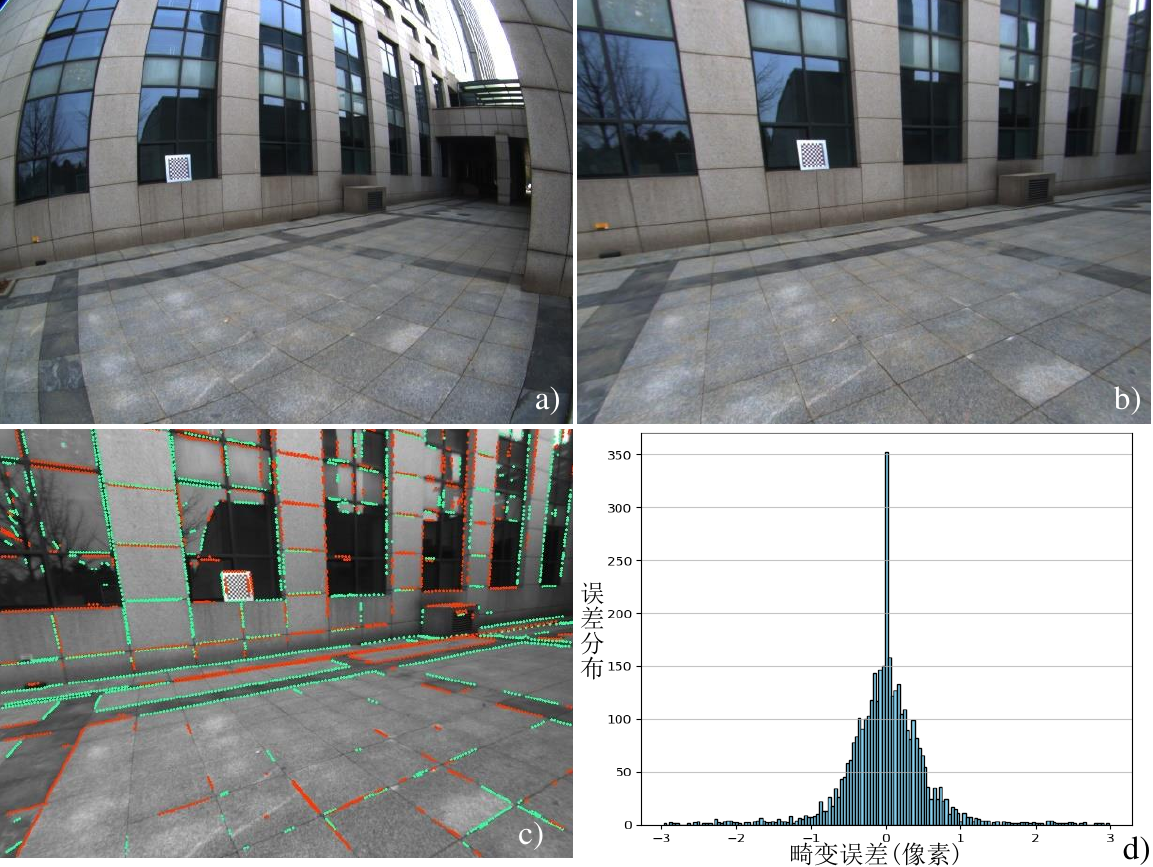}
    \caption{(a) is the original image. (b) is the image after undistortion using the calibrated intrinsic parameters. (c) is the extraction of line segments from the undistorted image. (d) is the error analysis of the undistorted image.}
    \label{fig:intrinsics_experiment}
\end{figure}

To quantitatively assess the accuracy of camera intrinsics, this paper employs the calibration harp \cite{calib_harp} to evaluate the camera's distortion. The scene used for the intrinsics experiment is shown in Fig. \ref{fig:intrinsics_experiment} (a), which consists of many parallel lines. First, a series of photos of the "calibration harp" are taken with the camera. Then, the images are undistorted using calibration parameters to obtain corrected images, as shown in Fig. \ref{fig:intrinsics_experiment} (b). Subsequently, the LSD algorithm is used to detect line segments in the images, which correspond to the straight lines on the "calibration harp," as shown in Fig .\ref{fig:intrinsics_experiment} (c). In the figure, 383 lines are detected. The detected edge points are resampled and processed with Gaussian blurring to reduce noise and enhance the accuracy of the edge points.

By statistically sampling the distance of points to the line and removing outliers, as shown in Fig .\ref{fig:intrinsics_experiment} (d), it can be observed that the distortion error is within 0.5 pixels, the root mean square error is 1.075 pixels, and the maximum error distance is 2.152 pixels, thereby verifying the effectiveness of the algorithm.

\subsection{LiDAR-Camera Extrinsic Parameters Calibration Experiment}

\begin{figure}
    \vspace{0cm}
    \setlength{\abovecaptionskip}{0.1cm} 
    \centering
    \includegraphics[width = 8.5cm]{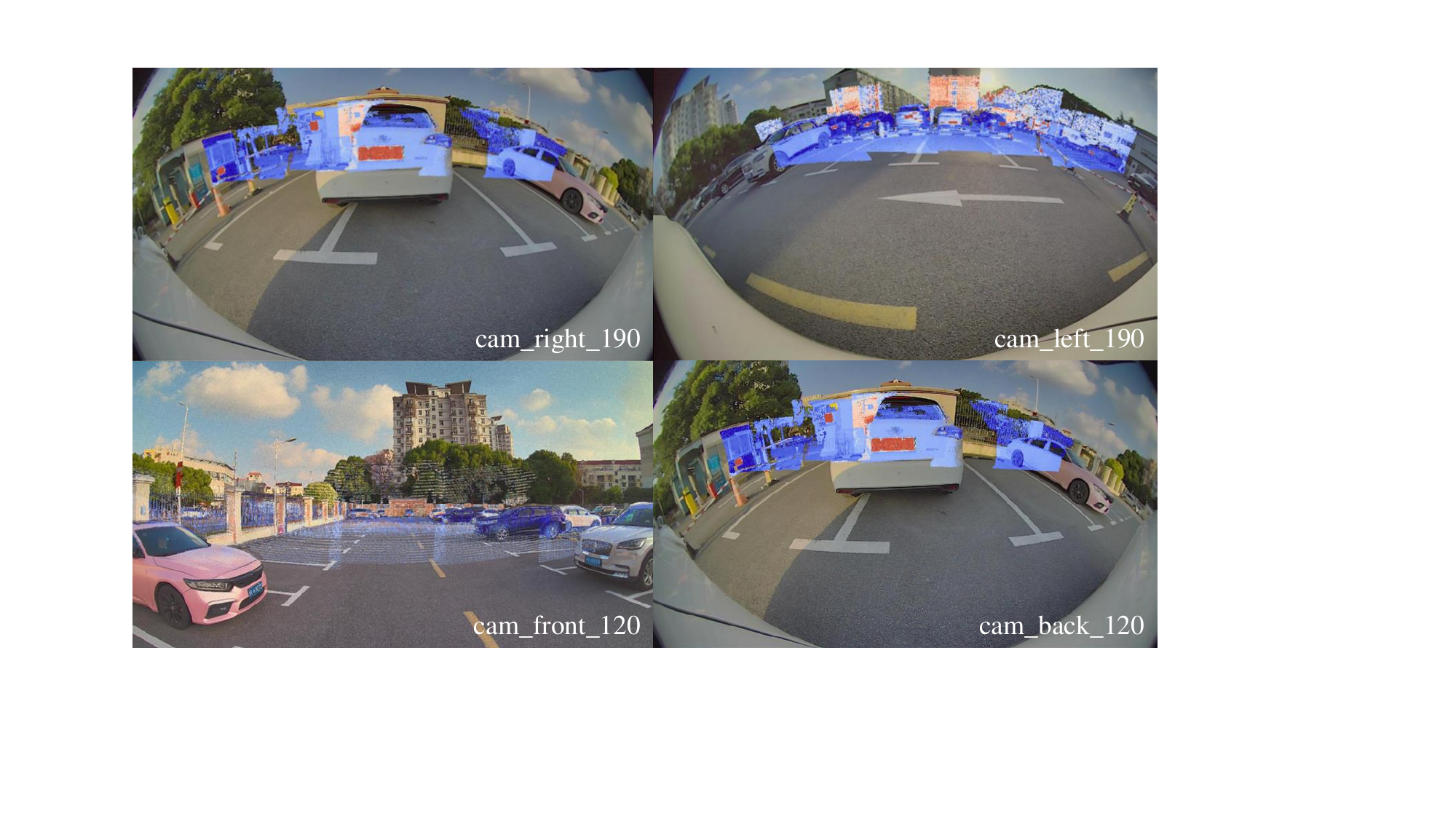}
    \caption{Based on the camera imaging model, project the LiDAR point cloud onto the image plane with different fields of view (FOV).}
    \label{fig:l2c_experiment}
\end{figure}

To validate the accuracy of the extrinsic parameters between the LiDAR and the camera, this paper selects static frame images of C\_F, C\_E, C\_D and C\_A. The LiDAR point cloud that shares a common viewing area with these images is projected onto the imaging plane according to the camera imaging model. The projection effect is shown in Fig. \ref{fig:l2c_experiment}. It can be observed that the parameters calibrated by the algorithm in this paper can simultaneously satisfy the co-viewing relationship between wide-angle and ultra-wide-angle cameras and the LiDAR.

\begin{table}[]
\small
\centering
\tabcolsep=0.1cm
\caption{Reprojection error of LiDAR projection to camera plane.}
\label{table:reprojection_error}
\setlength{\tabcolsep}{2pt}
\renewcommand{\arraystretch}{1.4}
\begin{tabular}{ccccc}
\hline
Error & L\_A\_to\_C\_A & L\_B\_to\_C\_D & L\_C\_to\_C\_E & L\_D\_to\_C\_F \\ \hline
ours               & 1.6132           & 1.8345           & 1.6376           & 1.7345           \\ \hline
\end{tabular}
\end{table}

\subsection{LiDAR-IMU Extrinsic Parameters Calibration Experiment}

\begin{figure}
    \vspace{0cm}
    \setlength{\abovecaptionskip}{0.1cm} 
    \centering
    \includegraphics[width = 8.5cm]{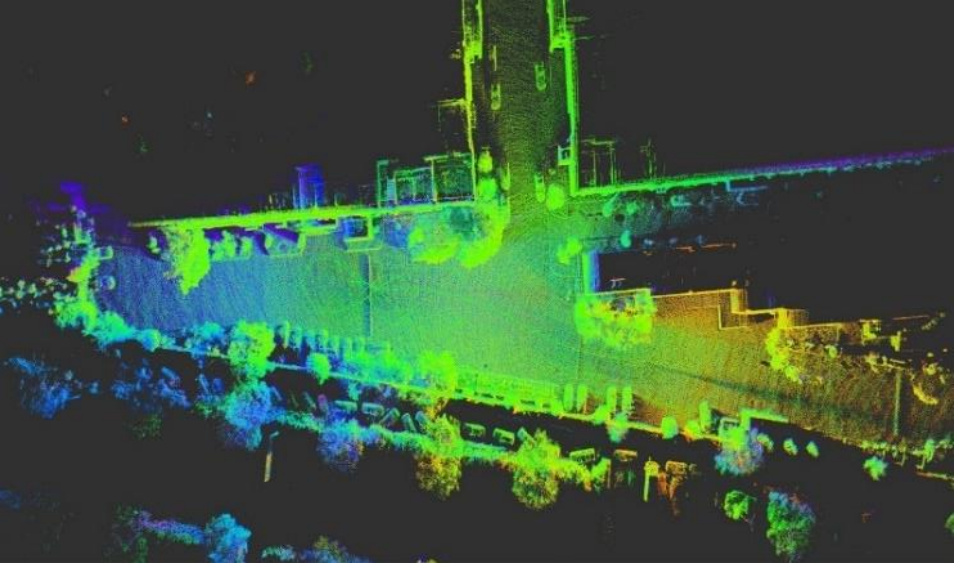}
    \caption{Based on the calibrated extrinsic parameters between the LiDAR and IMU, use the true pose provided by the Inertial Navigation System (INS) to deduce the LiDAR's point cloud map. The consistency of the point cloud map can be used to verify the effectiveness of the method presented in this paper.}
    \label{fig:l2i_experiment}
    \vspace{-0.3cm}
\end{figure}

Autonomous vehicles are equipped with high-precision inertial navigation RTK, which can provide centimeter-level pose truth for the IMU. Based on the IMU's pose and the extrinsic parameters between the LiDAR and IMU, the LiDAR point cloud map can be deduced. The stitching effect is shown in Fig. \ref{fig:l2i_experiment}. By measuring the accuracy of the deduced LiDAR point cloud map, the accuracy of the extrinsic parameters can be assessed. This paper adopts the Mean Map Entropy (MME) \cite{MME} as the map quality indicator.

To further verify that the algorithm presented in this paper is also applicable to the calibration of any number of LiDARs, the paper conducts experiments by dividing four LiDARs into three groups to validate the consistency of calibration, which are the front and rear LiDAR joint calibration experiment, the left and right LiDAR joint calibration, and the front, rear, left, and right LiDAR joint calibration. The calibration results are shown in Table \ref{table:LiDAR_Group}. The maximum angle error of the method proposed in this paper is 0.05°, and the maximum translation error is 2cm.

\begin{table*}[]
\small
\centering
\tabcolsep=0.1cm
\caption{Multi-LiDAR Group Joint Calibration Accuracy Assessment.}
\label{table:LiDAR_Group}
\setlength{\tabcolsep}{3.5pt}
\renewcommand{\arraystretch}{1.3}
\begin{tabular}{ccccc}
\hline
calib result                                                                          &       & L\_A with L\_B     & L\_C with L\_D    & All LiDAR           \\ \hline
\multirow{4}{*}{\begin{tabular}[c]{@{}c@{}}rotation\\ (roll pitch yaw°)\end{tabular}} & L\_B & 99.13 89.18 9.13     & \textbackslash{}    & 99.02 89.16 9.03    \\
                                                                                      & L\_A & -90.164 -4.87 179.13 & \textbackslash{}    & -90.16 -4.83 179.2  \\
                                                                                      & L\_D & \textbackslash{}     & 179.49 -6.31 179.80 & 179.54 -6.31 179.81 \\
                                                                                      & L\_C & \textbackslash{}     & -0.81 5.08 -179.51  & -0.86 -4.99 -179.46 \\ \hline
\multirow{4}{*}{\begin{tabular}[c]{@{}c@{}}translation\\ (tx ty m)\end{tabular}}      & L\_B & 0.032 1.697          & \textbackslash{}    & 0.012 1.693         \\
                                                                                      & L\_A & -0.013 0.13          & \textbackslash{}    & 0.007 0.130         \\
                                                                                      & L\_D & \textbackslash{}     & -0.423 0.963        & -0.423 0.967        \\
                                                                                      & L\_C & \textbackslash{}     & 0.511 0.958         & 0.511 0.976         \\ \hline
\end{tabular}
\end{table*}

\begin{table}[]
\small
\centering
\tabcolsep=0.1cm
\caption{LiDAR map entropy.}
\label{table:map_entropy}
\setlength{\tabcolsep}{3.5pt}
\renewcommand{\arraystretch}{1.5}
\begin{tabular}{ccccc}
\hline
MME   & L\_B                      & L\_A                      & L\_C                      & \multicolumn{1}{c}{L\_D} \\ \hline
+0.5° & -5.186                     & -5.026                     & -5.149                     & -5.125                    \\
-0.5° & \multicolumn{1}{l}{-5.125} & \multicolumn{1}{l}{-5.054} & \multicolumn{1}{l}{-5.256} & -5.234                    \\
ours  & -5.395                     & -5.339                     & -5.454                     & -5.468                    \\ \hline
\end{tabular}
\end{table}

\subsection{LiDAR-LiDAR Extrinsic Parameters Calibration Experiment}

\begin{figure}
    \vspace{0cm}
    \setlength{\abovecaptionskip}{0.1cm} 
    \centering
    \includegraphics[width = 8.5cm]{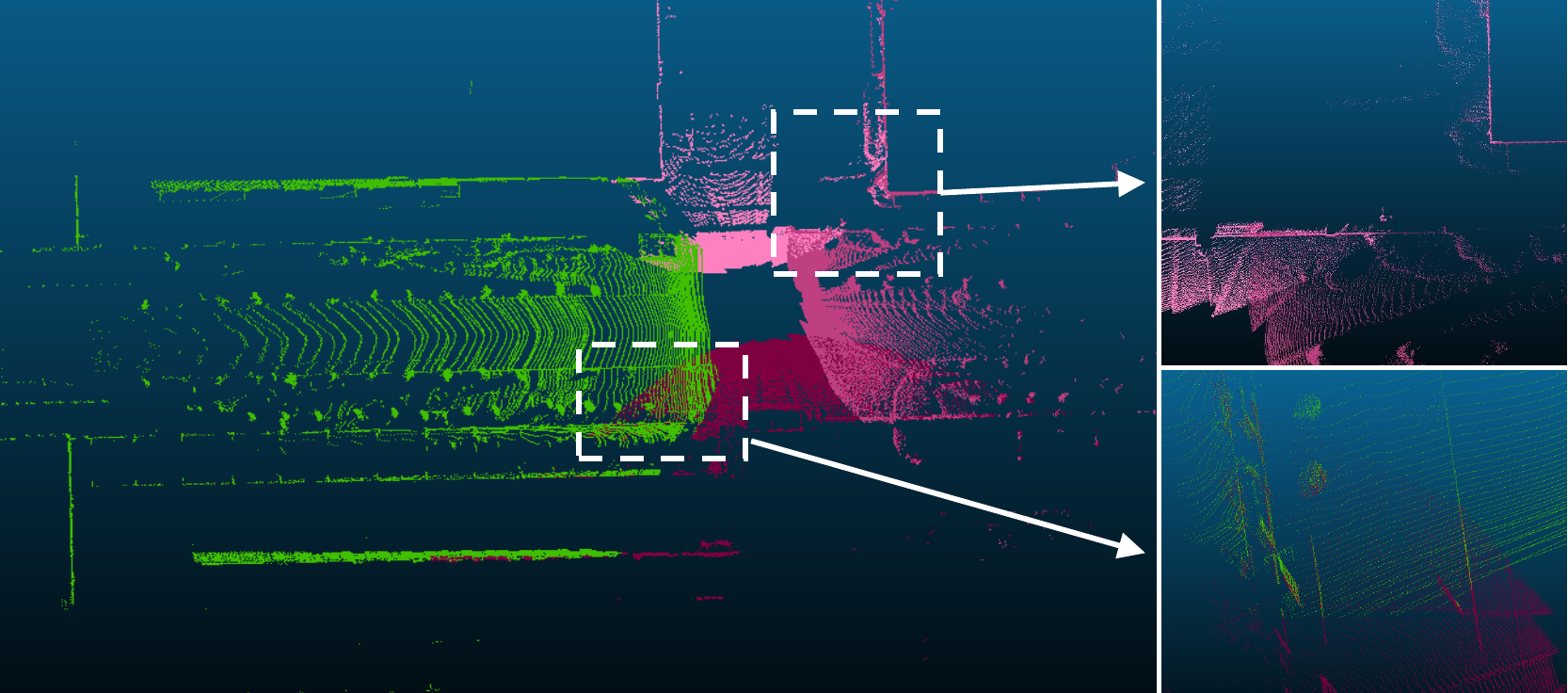}
    \caption{Merging LiDAR single-frame point clouds using the extrinsic parameters calibrated in this paper.}
    \label{fig:l2l_experiment}
    \vspace{-0.3cm}
\end{figure}

This paper selects static frame point clouds from four LiDARs and performs point cloud stitching using the extrinsic parameters calibrated by the method presented in this paper. Each LiDAR point cloud is assigned a different color, and the stitching effect is shown in Fig. \ref{fig:l2l_experiment}. From the stitching points between the LiDARs in the figure, it can be seen that the point clouds from different LiDARs are essentially overlapping. This is because this paper has constructed the co-viewing relationship between the LiDARs based on an adaptive voxel map, resulting in a high degree of overlap between the LiDARs.

To further verify the extrinsic parameters between the LiDARs, we extracts the point clouds within the common viewing area between different LiDARs after stitching and calculates the thickness of the point clouds. The thickness of the point clouds is shown in Table \ref{table:Multi-LiDAR}. It can be seen that the thickness of the point clouds is within 2cm. Since the distance measurement error of the LiDAR is approximately 2cm, the extrinsic parameter error calibrated by the method in this paper is within a reasonable range. The method described in this paper is compared with hand-eye calibration, as the latter does not establish data associations between the LiDARs, resulting in a weaker overlap between the LiDARs than the method presented in this paper.

\begin{table}[]
\small
\centering
\tabcolsep=0.1cm
\caption{The accuracy of Multi-LiDAR calibration is compared with that of hand-eye calibration. (CM)}
\label{table:Multi-LiDAR}
\setlength{\tabcolsep}{3.5pt}
\renewcommand{\arraystretch}{1.3}
\begin{tabular}{ccccc}
\hline
Thickness & L\_B\_to\_L\_C & L\_C\_to\_L\_A & L\_A\_to\_L\_D & L\_D\_to\_L\_B \\ \hline
ours                  & 2.285      & 2.207      & 1.867      & 1.902      \\
Hand-eye  & 4.348      & 4.556      & 5.424        & 4.254        \\ \hline
\end{tabular}
\end{table}

\subsection{Camera-Camera Extrinsic Parameters Calibration Experiment}

This paper uses the epipolar error to measure the accuracy of the camera parameters, calculating the fundamental matrix between two frames of images using the intrinsic and extrinsic parameters obtained from calibration. Corresponding feature points are detected and matched in the images of the two views. For each pair of matched points, the distance to their respective epipolar lines is calculated based on the fundamental matrix. It reflects the deviation between the geometric position of the matched points and the ideal epipolar geometric constraints. By statistically analyzing the distribution of the epipolar errors, as shown in Table \ref{table:Multi-Camera}, the epipolar error is within 1.5 pixels, and the low value of the epipolar error implies a higher calibration accuracy. Compared to methods without hand-eye calibration, the method presented in this paper has higher accuracy.

\begin{table}[]
\small
\centering
\tabcolsep=0.1cm
\caption{The method presented in this paper is compared with Multi-Camera calibration accuracy of hand-eye calibration. (pixel)}
\label{table:Multi-Camera}
\setlength{\tabcolsep}{3.5pt}
\renewcommand{\arraystretch}{1.3}
\begin{tabular}{ccccc}
\hline
Epipolar Error & C\_D\_to\_C\_E & C\_D\_to\_C\_F & C\_E\_to\_C\_F \\ \hline
ours           & 1.361        & 1.29         & 0.89         \\
Hand-eye       & 3.564        & 3.985        & 3.647        \\ \hline
\end{tabular}
\end{table}

\begin{figure}
    \vspace{0cm}
    \setlength{\abovecaptionskip}{0.1cm} 
    \centering
    \includegraphics[width = 8.5cm]{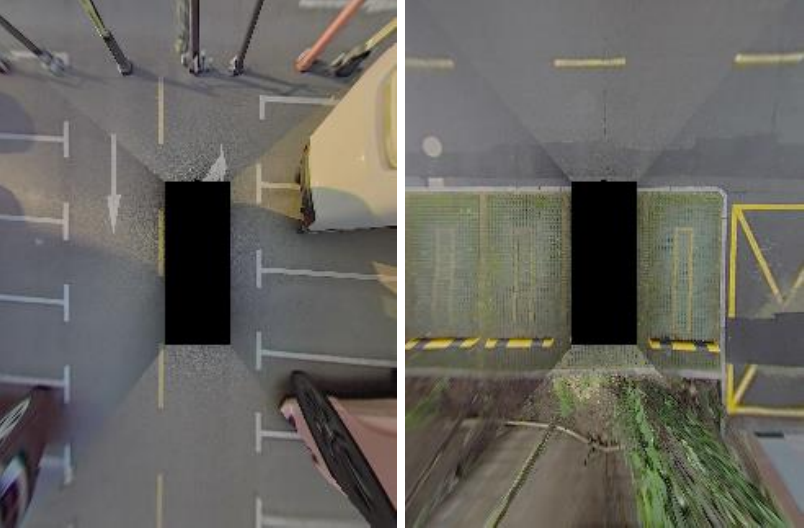}
    \caption{Merging the images of FOV190 from a Bird's Eye View (BEV) perspective using the extrinsic parameters calibrated in this paper.}
    \label{fig:c2c_bev}
\end{figure}

To further verify the accuracy of the camera extrinsic calibration, this paper converts the FOV190 static frame images used for parking from the camera coordinate system to the ground coordinate system, and performs image stitching from a bird's eye view (BEV) perspective to test the accuracy of the extrinsic calibration. Four FOV190 images are transformed into the ground coordinate system for stitching, and the stitching effect is shown in Fig. \ref{fig:c2c_bev}. From the stitching seam, it can be seen that the image stitching effect is quite good.

\section{CONCLUSIONS}

Facing the challenges of sensor asynchrony and non-overlapping fields of view among devices, this paper proposes a continuous-time-based joint calibration algorithm. Utilizing the spatial structure and texture information in natural scenes to build data associations between sensors, the algorithm performs joint calibration of all sensors' co-viewing features and motion information within a continuous time framework. Initially, it conducts self-calibration of camera intrinsics through Structure from Motion (SFM) and establishes data associations between cameras. Subsequently, it constructs data associations between LiDARs using an adaptive voxel map and optimizes the map for extrinsic calibration. Finally, by feature matching, it creates data associations between the intensity images of the LiDAR map projections and the camera images, achieving joint optimization of intrinsic and extrinsic parameters. This method is capable of jointly calibrating any number of cameras and LiDARs, thus avoiding the cumulative errors that arise from separate calibrations.

\bibliographystyle{IEEEtran}
\bibliography{ref}

\vfill

\end{document}